%% file: arxiv_TreeMAML.tex
\documentclass[11pt,a4paper]{article}

\usepackage{authblk}

\usepackage[hyperref]{eacl2021}
\usepackage{times}
\usepackage{latexsym}

\input{math_commands.tex}

\usepackage{microtype}

\aclfinalcopy 

\usepackage{algorithm}
\usepackage{algorithmic}
\usepackage{adjustbox}

\title{Meta-Learning with \\
            MAML on Trees}

\author[*]{Jezabel R. Garcia, Federica Freddi, Feng-Ting Liao, Jamie McGowan, Tim Nieradzik,\\Da-shan Shiu, Ye Tian, Alberto Bernacchia}
\affil[*]{MediaTek Research, Cambourne Business Park, Cambridge CB23 6DW, United Kingdom}

\date{}

\begin{document}
\maketitle
\begin{abstract}
In meta-learning, the knowledge learned from previous tasks is transferred to new ones, but this transfer only works if tasks are related. Sharing information between unrelated tasks might hurt performance, and it is unclear how to transfer knowledge across tasks with a hierarchical structure. Our research extends a  model agnostic meta-learning model, MAML, by exploiting hierarchical task relationships. Our algorithm, TreeMAML, adapts the model to each task with a few gradient steps, but the adaptation follows the hierarchical tree structure: in each step, gradients are pooled across tasks clusters, and subsequent steps follow down the tree. We also implement a clustering algorithm that generates the tasks tree without previous knowledge of the task structure, allowing us to make use of implicit relationships between the tasks.
We show that the new algorithm, which we term TreeMAML, performs better than MAML when the task structure is hierarchical for synthetic experiments. To study the performance of the method in real-world data, we apply this method to Natural Language Understanding, we use our algorithm to finetune Language Models taking advantage of the language phylogenetic tree.  We show that TreeMAML improves the state of the art results for cross-lingual Natural Language Inference. This result is useful, since most languages in the world are under-resourced and the improvement on cross-lingual transfer allows the internationalization of NLP models.
   This results open the window to use this algorithm in other  real-world hierarchical  datasets.  
   
\end{abstract}

\section{Introduction}

Deep learning models require a large amount of data in order to perform well when trained from scratch.
When data is scarce for a given task, we can \emph{transfer} the knowledge gained in a source task to quickly learn a target task, if the two tasks are related. 
\emph{Multi-task learning} studies how to learn multiple tasks simultaneously with a single model, by taking advantage of task relationships \cite{ruder_overview_2017, zhang_survey_2018}.
However, in Multi-task learning models, a set of tasks is fixed in advance and they do not generalize to new tasks.
Instead, \emph{Meta-learning} is inspired by the human ability to learn how to quickly learn new tasks by using the knowledge of previously learned ones.

Meta-learning has been widely used in multiple domains, especially in recent years since the advent of Deep Learning \cite{hospedales_meta-learning_2020}. 
A successful model for meta-learning, MAML \cite{finn_model-agnostic_2017}, does not diversify task relationships according to their similarity and it is unclear how to modify it for that purpose.
Furthermore, there is still a lack of methods for sharing information across tasks that have a hierarchical structure, and the goal of our work is to fill this gap.

The use of MAML-like algorithms in NLP has just recently been proved successful for Natural Language Inference (NLI) and Question Answering (QA) \cite{nooralahzadeh_zero-shot_2020}. These results represent a practical meta-learning solution to the fundamental problem of applying NLP models to under-resourced languages where data annotation is scarce. This work, combined with the fact that languages can be organized hierarchically using their phylogenetic tree \cite{dunn_evolved_2011}, motivated us to develop a hierarchical meta-learning algorithm, that we call TreeMAML.

In this work, we make the following contributions:
\begin{itemize}

    \item We propose a novel modification of MAML to account for a hierarchy of tasks. The algorithm uses the tree structure of data during adaptation, by pooling gradients across tasks at each adaptation step and subsequent steps follow down the tree (see Figure 1a).
    
    \item We modify the hierarchical clustering from \citet{menon_online_2019} to allow asymmetric tree structure. We apply this clustering algorithm to learn dynamic trees that exploit the similarity between tasks.
    
     \item We introduce new benchmarks for testing a hierarchy of tasks in meta-learning using a multidimensional linear regression task. We compare our algorithm to MAML and a baseline model, where we train on all tasks without any meta-learning algorithm applied.
    
    \item  We apply TreeMAML to few-shot NLI, using the XNLI dataset \cite{conneau_xnli_2018}, obtaining accuracies higher than previous state-of-the-art.
    
\end{itemize}

\section{Related work}

The problem of quantifying and exploiting task relationships has a long history in Multi-task learning and is usually approached by parameter sharing, see \citet{ruder_overview_2017,zhang_survey_2018} for reviews.
However, Multi-task Learning is fundamentally different from Meta-learning as it does not consider the problem of generalizing to new tasks \cite{hospedales_meta-learning_2020}.
Recent work includes \citet{zamir_taskonomy_2018}, who studies a large number of computer vision tasks and quantifies the transfer between all pairs of tasks.
\citet{achille_task2vec_2019} proposes a novel measure of task representation by assigning an importance score to each model parameter in each task.
The score is based on each task's loss function gradients with respect to each model parameter.
This work suggests that gradients can be used as a measure of task similarity and we use this insight in our proposed algorithm.

In Meta-learning, a few papers have been recently published on learning and using task relationships.
The work  of \citet{yao_hierarchically_2019} applies hierarchical clustering to task representations learned by an autoencoder and uses those clusters to adapt the parameters to each task.
The model of \citet{liu_learning_2019} maps the classes of each task into the edges of a graph, it meta-learns relationships between classes and how to allocate new classes by using a graph neural network with attention.
However, these algorithms are not model-agnostic; they have a fixed backbone and loss function and are thus difficult to apply to new problems.
Instead, we design our algorithm as a straightforward generalization of Model-agnostic meta-learning (MAML, \citet{finn_model-agnostic_2017}) and it can be applied to any loss function and backbone.

A couple of studies looked into modifying MAML to account for task similarities.
The work of \citet{jerfel_reconciling_2019} finds a different initial condition for each cluster of tasks and applies the algorithm to the problem of continual learning.
The work of \citet{katoch_invenio_2020} defines parameter updates for a task by aggregating gradients from other tasks according to their similarity.
However, in contrast with our algorithm, both of these models are not hierarchical, tasks are clustered on one level only and cannot be represented by a tree structure.

Recently, MAML has been applied to cross-lingual meta-learning \cite{gu_meta-learning_2018, dou_investigating_2019}. In particular, the implementation by \citet{nooralahzadeh_zero-shot_2020}, called XMAML, obtained good results on NLI and QA tasks. As in the previously mentioned computer vision studies, some of these NLP algorithms looked into the relationships among languages to select the support languages used in their meta-learning algorithm, but they do not use the hierarchical structure of the languages. 

\section{The meta-learning problem}

We follow the notation of \citet{hospedales_meta-learning_2020}.
We assume the existence of a distribution over tasks $\tau$ and, for each task, a distribution over data points $\mathcal{D}$ and a loss function $\mathcal{L}$.
The loss function of the meta-learning problem, $\mathcal{L}^{meta}$, is defined as an average across both distributions of tasks and data points:
\begin{equation}
\mathcal{L}^{meta}\left(\boldsymbol\omega\right)=\mathop{\mathbb{E}}_{\tau}\mathop{\mathbb{E}}_{\mathcal{D}|\tau}\mathcal{L}_{\tau}\left(\boldsymbol\theta_{\tau}(\boldsymbol\omega);\mathcal{D}\right)
\end{equation}
The goal of meta-learning is to minimize the loss function with respect to a vector of meta-parameters $\boldsymbol\omega$.
The vector of parameters $\boldsymbol\theta$ is task-specific and depends on the meta-parameters $\boldsymbol\omega$.
Different meta-learning algorithms correspond to a different choice of $\boldsymbol\theta_{\tau}(\boldsymbol\omega)$.
We describe below the choice of MAML that will also be followed by TreeMAML.

During meta-training, the loss is evaluated on a sample of $m$ tasks and  $n_v$ validation data points for each task.
\begin{equation}
\label{loss}
\mathcal{L}^{meta}\left(\boldsymbol\omega\right)=\frac{1}{mn_v}\sum_{i=1}^m\sum_{j=1}^{n_v}\mathcal{L}_{\tau_i}\left(\boldsymbol\theta_{\tau_i}(\boldsymbol\omega);\mathcal{D}_{ij}\right)
\end{equation}
For each task $i$, the parameters $\boldsymbol\theta_{\tau_i}$ are learned by a set of $n_t$ training data points, distinct from the validation data.
During meta-testing, a new (target) task is given and the parameters $\boldsymbol\theta$ are learned by a set of $n_r$ target data points.
In this work, we also use a batch of training data points to adapt $\boldsymbol\theta$ at test time.
No training data is used to compute the model's final performance, which is computed on separate test data of the target task.

\subsection{TreeMAML}
MAML aims at finding the optimal initial condition $\omega$ from which a suitable parameter set can be found, separately for each task, after $K$ gradient steps \cite{finn_model-agnostic_2017}.
For task $i$, we define the single gradient step with learning rate $\alpha$ as
\begin{equation}
U_i(\boldsymbol\omega)=\boldsymbol\omega-\frac{\alpha}{n_t}\sum_{j=1}^{n_t}\nabla \mathcal{L}(\boldsymbol\omega;\mathcal{D}_{ij})
\end{equation}
Then, MAML with $K$ gradient steps corresponds to $K$ iterations of this step.
\begin{equation}
\boldsymbol\theta_{\tau_i}(\boldsymbol\omega)=U_i(U_i(...U_i(\boldsymbol\omega)))\;\;\;\;\;\;\;\;\;\;\;\;\mbox{($K$ times)}
\end{equation}
This update is usually referred to as \emph{inner loop} and is performed separately for each task, while optimization of the loss \ref{loss} is referred to as \emph{outer loop}.

\begin{figure*}[h]
\begin{center}
\includegraphics[width=9cm]{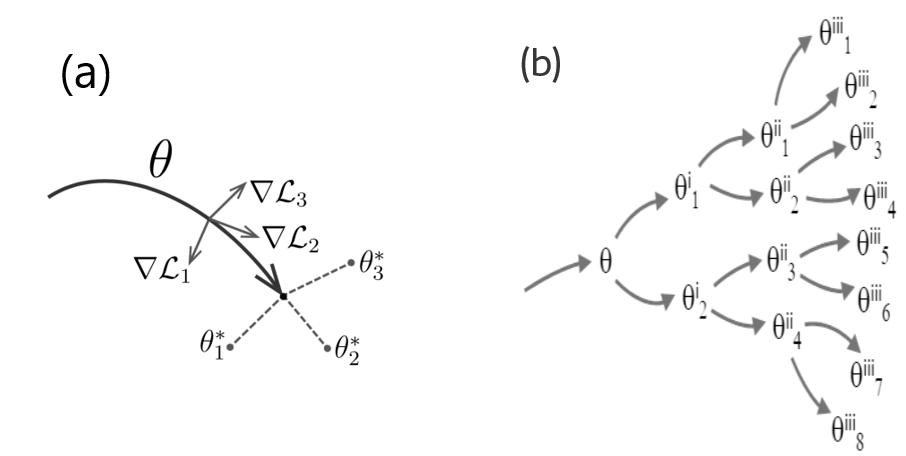}

\end{center}
\caption{Illustration of the MAML(a) and TreeMAML(b) algorithms. Both algorithms are designed to quickly adapt to new tasks with a small number of training samples. MAML achieves this by introducing a gradient step in the direction of the single task. TreeMAML follows a similar approach, but it exploits the relationship between tasks by introducing the hierarchical aggregation of the gradients.}
\label{fig1}
\end{figure*}

We propose to modify MAML in order to account for a hierarchical structure of tasks.
The idea is illustrated in Figure \ref{fig1}.

At each gradient step $k$, we assume that tasks are aggregated into $C_k$ clusters and the parameters for each task are updated according to the average gradient across tasks within the corresponding cluster (in Fig.1b, we use $K=3$ steps and $C_1=2$, $C_2=4$, $C_3=8$).
We denote by $\mathcal{T}_c$ the set of tasks in cluster $c$.
Then, the gradient update for the parameters of each task belonging to cluster $c$ is equal to
\begin{equation}
U_c(\boldsymbol\omega)=\boldsymbol\omega-\frac{\alpha}{n_t\left|\mathcal{T}_c\right|} \sum_{i\in \mathcal{T}_c}\sum_{j=1}^{n_t}\nabla \mathcal{L}(\boldsymbol\omega;\mathcal{D}^{(i)}_j)
\end{equation}
Furthermore, we denote by $c_i^k$ the cluster to which task $i$ belongs at step $k$.
Then, TreeMAML with $k$ gradient steps corresponds to $K$ iterations of this step.
\begin{equation}
\label{endinner}
\boldsymbol\theta_{\tau_i}(\boldsymbol\omega)=U_{c_i^K}(U_{c_i^{K-1}}(...U_{c_i^1}(\boldsymbol\omega)))
\end{equation}
The intuition is the following: if each task has scarce data, gradient updates for single tasks are noisy and adding up gradients across similar tasks increases the signal.
Note that we recover MAML if $C_k$ is equal to the total number of tasks $m$ at all steps.
On the other hand, if $C_k=1$, then the inner loop would take a step with a gradient averaged across all tasks.

Because at one specific step the weight updates are equal for all tasks within a cluster, it is possible to define the steps of the inner loop update per cluster $c$ instead of per task $\boldsymbol\theta_{\tau_i}$.
Given a cluster $c$ and its parent cluster $p_c$ in the tree, the update at step $k$ is given by
\begin{equation}
\boldsymbol\theta_{c,k}=\boldsymbol\theta_{p_c,k-1}-\frac{\alpha}{n_t\left|\mathcal{T}_c\right|} \sum_{i\in \mathcal{T}_c}\sum_{j=1}^{n_t}\nabla \mathcal{L}(\boldsymbol\theta_{p_c,k-1};\mathcal{D}_{ij})
\end{equation}
where $\boldsymbol\theta^{c}_k$ is the parameter value for cluster $c$ at step $k$.
In terms of the notation used in expression \ref{endinner}, we have the equivalence $\boldsymbol\theta_{\tau_i}(\boldsymbol\omega)=\boldsymbol\theta_{c_i,K}$, which depends on the initial condition $\boldsymbol\omega$.
The full procedure is described in Algorithm \ref{alg:treemaml_alg}

We consider two versions of the algorithm, depending on how we obtain the tree structure similar to \citet{srivastava_discriminative_2013}:
\begin{itemize}
    \item \textbf{Fixed TreeMAML}. The tree is fixed by the knowledge of the tree structure of tasks when this structure is available. In that case, the values of $C_k$ are determined by such tree.
    \item \textbf{Learned TreeMAML}. The tree is unknown \emph{a priori} and is learned using a hierarchical clustering algorithm. In that case, the values of $C_k$ are determined at each step by the clustering algorithm.
\end{itemize}

In the latter case, we cluster tasks based on the gradients of each task loss, consistent with recent work \citet{achille_task2vec_2019}.
After each step $k$ at cluster $c_i$, the clustering algorithm takes as input the gradient vectors of the children tasks $i$
\begin{equation}
\mathbf{g}_{ik}=\frac{1}{n_t}\sum_{j=1}^{n_t}\nabla \mathcal{L}(\boldsymbol\theta_{c_i,k};\mathcal{D}_{ij}) \end{equation}
and these gradients are further allocated into clusters according to their similarity.
The clustering algorithm is described in subsection \ref{cluster}.

Similar to MAML, adaptation to a new task is performed by computing $\boldsymbol\theta^{(i)}(\boldsymbol\omega)$ on a batch of data of the target task. In order to exploit task relationships, we first reconstruct the tree structure by using a batch of training data and then we introduce the new task.

\begin{algorithm*}[h!]
\caption{TreeMAML}
\begin{algorithmic}
 \label{alg:treemaml_alg}
\REQUIRE distribution over tasks $p(\tau)$; distribution over data for each task $p(\mathcal{D}|\tau)$;  
\REQUIRE number of inner steps $K$; number of training tasks $m$; learning rates $\alpha, \beta$;
\REQUIRE number of clusters $C_k$ for each step $k$; loss function $\mathcal{L}_{\tau}(\boldsymbol\omega,\mathcal{D})$ for each task
\STATE randomly initialize $\boldsymbol\omega$
\WHILE{not done \do}
\STATE sample batch of $i=1:m$ tasks $\{\tau_{i}\}\sim p(\tau)$ 
    \STATE for all tasks $i=1:m$ initialize a single cluster $c_i=1$
    \STATE initialize $\boldsymbol\theta_{1,0}=\boldsymbol\omega$
\FOR {steps $k=1:K$}
\FOR{tasks $i=1:m$}
    \STATE sample batch of $j=1:n_v$ data points $\{\mathcal{D}_{ij}\}\sim p(\mathcal{D}|\tau_i)$
    \STATE evaluate gradient $\mathbf{g}_{ik}=\frac{1}{n_t}\sum_{j=1}^{n_t}\nabla \mathcal{L}_{\tau_i}(\boldsymbol\theta_{c_i,k-1};\mathcal{D}_{ij})$
\ENDFOR
    \STATE regroup tasks into $C_{k}$ clusters $\mathcal{T}_{c}=\{i:c_i=c\}$  
    \STATE according to similarity of $\{\mathbf{g}_{ik}\}$ and parent clusters $\{p_c\}$ 
    \STATE update $\theta_{c,k}=\theta_{p_c,k-1}-\frac{\alpha}{|\mathcal{T}_{c}|}\sum_{i\in \mathcal{T}_{c}}\mathbf{g}_{ik}$ for all clusters $c=1:C_k$
\ENDFOR
\STATE update $\boldsymbol\omega \leftarrow\boldsymbol\omega-\beta\frac{1}{mn_v}\sum_{i=1}^m\sum_{j=1}^{n_v}\nabla_{\boldsymbol\omega}\mathcal{L}_{\tau_i}\left(\boldsymbol\theta_{c_i,K}(\boldsymbol\omega);\mathcal{D}_{ij}\right)$ 
\ENDWHILE
\end{algorithmic}
\end{algorithm*}

\subsection{Clustering Algorithm} \label{cluster}
We employ a hierarchical clustering algorithm to cluster the gradients of our model parameters in the learned TreeMAML case.
We specifically opt for an online clustering algorithm to maximise computational efficiency at test time and scalability. When a new task is evaluated, we reuse the tree structure generated for a training batch and add the new task. This process saves us from computing a new task hierarchy from scratch for every new task.
Moreover, with offline hierarchical clustering, all the data needs to be available to the clustering algorithm simultaneously, which becomes a problem when dealing with larger batch sizes. Therefore online clustering favours scalability.

We follow the online top-down (OTD) approach set out by \citet{menon_online_2019} and adapt this to approximate non-binary tree structures. Our clustering algorithm is shown in Algorithm \ref{alg:cluster_alg}. Specifically, we introduce two modifications to the original OTD algorithm:
\begin{itemize}
    \item \emph{Maximum Tree Depth Parameter $D$}: This is equivalent to the number of inner steps to take in the TreeMAML since the tree is a representation of the inner loop where each layer in the tree represents a single inner step.
    \item \emph{Non-binary Tree Approximation}: We introduce a hyperparameter $\xi$ which represents how far the similarity of a new task needs to be to the average cluster similarity in order to be considered a child of that same cluster. This is not an absolute value of distance, but it is a multiplicative factor of the standard deviation of the intracluster similarities. Introducing this factor allows clusters at any level to have more than two children.
\end{itemize}
\begin{algorithm}[h!]
\caption{Online top down (OTD) - Non-binary}
\begin{algorithmic} 
\label{alg:cluster_alg}
\REQUIRE origin cluster node $C$ with a given set of children $A = \{x_1, x_2, ..x_N \}$
\REQUIRE new task $x$; maximum depth allowed $D$; similarity metric, $\omega()$
\REQUIRE standard deviation multiplicative hyperparameter $\xi$;
\IF { $|A|=0$}
    \STATE new task becomes a new child $A$ = $\{x\}$
\ELSIF {$|A|=1$}
    \STATE add new task to set of children $A \leftarrow  A \cup \{x\}$

\ELSIF {$\omega(A \cup \{x\}) > \omega(A)$}
    \STATE identify most similar child $x_* = \argmin_{x_i}(\omega(\{x_i, x\}))$ 
    \IF {reached maximum depth $\mathrm{C}_{\mathrm{depth}} + 1 = D$} 
        \STATE add new task to set of children $A \leftarrow A \cup \{x\}$ 
    \ELSE
        \STATE recursively perform OTD to create new node $C' = \textrm{OTD}(x_* ,x)$
        \STATE add new node to set of children $A \leftarrow ( A \setminus \{x_*\}) \cup {C'}$
    \ENDIF
\ELSIF {$\omega(A \cup \{x\}) < \omega(A) - \xi \sigma _{T}$}
    \STATE current node and new task become children to new cluster $A \leftarrow \{C, x\}$ 
\ELSE
    \STATE add new task to set of children $A \leftarrow A \cup \{x\}$
\ENDIF
\end{algorithmic}
\end{algorithm}

\section{Synthetic experiment}
In this section, we introduce a toy experiment illustrates the behaviour of fixed and learnt TreeMAML in a simple scenario.
We consider a multidimensional linear regression problem $y = \sum_{i=1}^{64} P_{i} x_{i} + \eta$ where the tasks are randomly sampled from a set of 4 defined clusters of multidimensional parameters $P$. These clusters of multi-dimensional parameters are selected to simulate hierarchically structured data.  $\eta$ is randomly generated Gaussian noise. Even in this case, the parameter clusters are arranged hierarchically such that $C_{1} = 2$, $C_{2} = 4$. 

The data points for the tasks are sampled uniformly $x_{i} \sim U[-5.0, 5.0]$ for all training and testing tasks where. The models are then trained and tested on a set of tasks with $K = 4, 8, 16, 32, 64$ and $128$ data points. 

\subsection{Fixed TreeMAML}
In these experiments, we assume knowledge about the structure of the underlying tasks and we use this to aggregate the gradients. 
\begin{figure}[ht] \centering
 \includegraphics[width=0.5\textwidth]{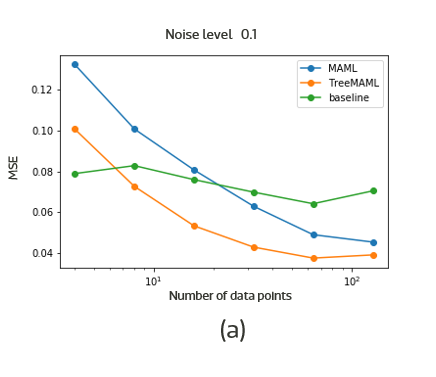}
  \includegraphics[width=0.5\textwidth]{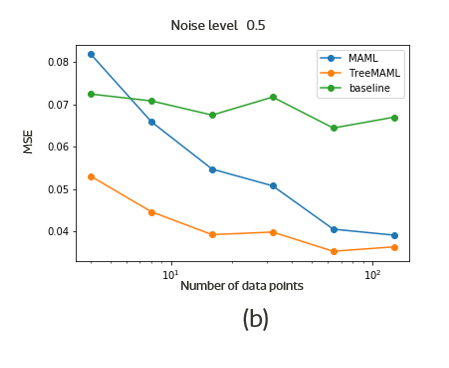}
  \caption{\label{fig:linear_result} Results of the multidimensional (N=64) linear regression task for Fixed TreeMAML, MAML and baseline for varying number of task data points $K = 4, 8, 16, 32, 64$ and $128$.}
\end{figure}
TreeMAML outperforms  MAML, especially when the number of tasks data points K is low. 

\subsection{Learned TreeMAML}
In this section, we assume no prior knowledge of the underlying structure of the data. Therefore, the data hierarchy is learnt per-batch using the modified OTD algorithm described in section \ref{cluster}. In the clustering algorithm, we set the maximum depth to 2 and we use the cosine similarity metric.  For this setting, the TreeMAML algorithm will perform three inner steps, where the last one is task-specific. Therefore, in order to make a fair comparison, in this experiment MAML is also set to perform three inner steps.

Table \ref{tab: lowNoiseRegResults} shows that TreeMAML outperforms the Baseline and MAML across all numbers of data points. What is more, learned TreeMAML performs better than the fixed tree for a larger number of data points. This is an expected effect since, as the number of data points increases, the gradients used to cluster the tasks will be less affected by the noise and become more accurate, leading to better clustering.
\begin{table*}
\begin{center}
\begin{tabular}{c|c|c|c}
    \hline
     Model & K=5 & k=10 & k=20 \\
     \hline
     Baseline & $1.293 \pm 0.074$ & $1.055 \pm 0.047$ & $1.139 \pm 0.061$ \\
     MAML & $1.025 \pm 0.068$ & $0.950 \pm 0.048$ & $0.785 \pm 0.028$ \\
     \hline
     \textbf{Fixed TreeMAML (ours)} & \boldmath$0.621 \pm 0.038$ & \boldmath$0.535 \pm 0.024$ & \boldmath$0.483 \pm 0.016$ \\
     \textbf{Learned TreeMAML (ours)} & $0.758 \pm 0.047$ & \boldmath$0.510 \pm 0.024$ & \boldmath$0.495 \pm 0.018$
\end{tabular}
\caption{\label{tab: lowNoiseRegResults}  Loss (MSE)  $\pm 95\%$ confidence intervals on multidimensional linear regression task, averaged over 400 meta-testing tasks. The results are presented for varying numbers of K data points and a noise level of 0.01}
 \end{center}
\end{table*}

\section{Cross-Lingual NLI} \label{cross-lingual}
\begin{figure*}[ht] \centering
 \includegraphics[width=0.8\textwidth]{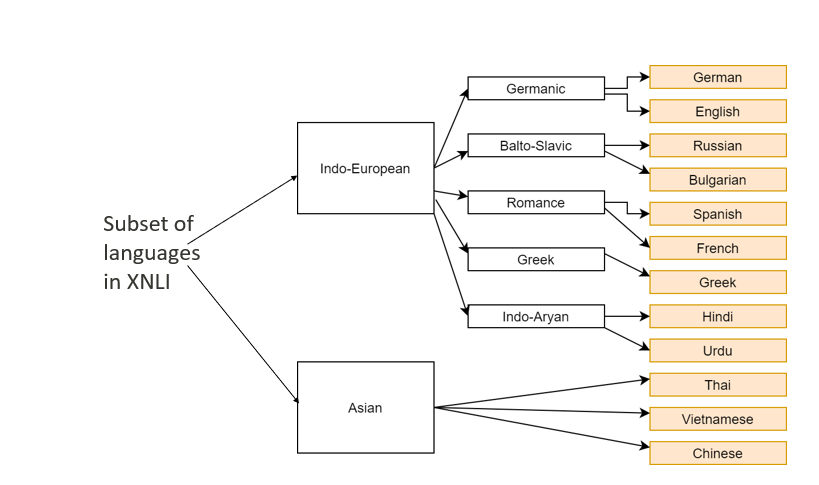}
  \caption{\label{fig:language_tree} Simplified version of the phylogenetic language tree. The tree include 12 of the 15 languages of XNLI data set and have depth three (Three levels of hierarchy) }
\end{figure*}

Languages can be embraced in a forest of phylogenetic trees \cite{dunn_evolved_2011}, for example, the Indo-European and generic Austric-Aisatic trees (Figure \ref{fig:language_tree}). TreeMAML exploits this hierarchical structure to generalize the performance of models across languages, including under-resourced languages, useing all the available languages in the tree.  

We adapt a high-resource language model, Multi-BERT \cite{devlin_bert_2018}, to a NLI task. In particular, we consider the problem of  
Few-Shot NLI using the XNLI data set \cite{conneau_xnli_2018}.

This dataset consists of a crowd-sourced collection of 5,000 test and 2,500 dev sentence-label pairs from the MultiNLI corpus. They are annotated with textual entailment and translated into 15 languages: English (en), French (fr), Spanish (es), German (de), Greek(el), Bulgarian (bg), Russian (ru), Turkish, Arabic, Vietnamese (vi), Thai (th), Chinese (zh), Hindi (hi), Swahili and Urdu (ur). Twelve of these languages are part of the same phylogenetic tree, and we focus our study on those languages (see Figure \ref{fig:language_tree}). 
We separately set as target language each language of the tree and we used the eleven remaining languages as auxiliary languages for meta-training.

Each sentence has also an associated topic, or genre, among a collection of $10$ possible genres (Face-To-Face, Telephone, Government, 9/11, Letters, Oxford University Press (OUP), Slate, Verbatim, and Government, Fiction).
We define each combination of a language and a genre as a task, and we consider the problem of few-shot meta-learning using three shots for each task during meta-training. 
 We add the new target task to the original distribution of tasks, we apply the TreeMAML algorithm and evaluate the model on the target language test set.

 We use the TreeMAML algorithm to fine-tune the top layer of Multi-BERT (layer 12), with four inner steps. We compare our results with MAML, using the same number of inner steps, with the baseline Multi-BERT, and with XMAML \citet{nooralahzadeh_zero-shot_2020}.      
An important difference of our approach is that, while XMAML uses only two auxiliary languages to fine tune Multi-BERT to a target language, we use all other languages as auxiliary languages.

\subsection{Fixed TreeMAML}
In the case of fixed TreeMAML, we use the phylogenetic tree in Figure \ref{fig:language_tree}. 
Fine-tuning of Multi-BERT for the target language benefits not only from proximal (auxiliary) languages, but also from all other languages in the tree that share roots with the target language. For example, if the target language is German, the fine-tuning in fixed TreeMAML would use for the first step of the gradient update all the remaining languages. In the second step, the auxiliary languages would be all the training languages of the Indo-European branch. The third and last steps uses only English. 
The accuracy of  TreeMAML is consistently higher than the one of the Baseline or MAML and an average of $\sim3\%$ better than the one achieved by XMAML, see Table \ref{tab: NLIResults}.

Note that we used a relatively simple version of the phylogenetic tree. A more detailed version could be used for testing under-resourced languages, or to emphasize the dependencies inside the tree. For example, a Bavarian testing data set would fall inside the German branch, or we could add depth to the tree by adding Germanic sub-branches, such as high-german, anglo-frisian and low-franconian.

\subsection{Learned TreeMAML} \label{treemaml-nli}
 While fixed TreeMAML uses previous knowledge to construct the tree, learned TreeMAML allows learning the relation between languages and genres, potentially reflecting a priory unknown relationships in the XNLI corpus, but also potentially fitting some noise.
 Note that learned TreeMAML has one additional parameter, the maximum tree depth, as explained in \ref{cluster}. 
 
 The relationships between languages and genres is learned at each step of gradient descent, for each batch of data. Therefore, the tree for one batch can be different from the tree for the next batch. This difference is due to the fact that the clustering algorithm only cares about the similarity of the gradients, and this similarity does not need to be always the same between two languages. It may depend on the particular words used in the sentences, or in the tasks genres. For example, for a particular batch, some sentences from the same genre in English and French could have closer gradients than other sentences in Germanic languages with a different genre. 
 
The clustering process happens at both training and testing time, which means that learned TreeMAML may improve the accuracy by improving the training, but also by producing the best hierarchy for the target task at testing time. This may be particularly useful for under-resourced languages where non-obvious dependencies between the task in the target languages and the tasks in other languages can be exploited to improve the test accuracy. 
\begin{table*}[ht]
\centering
\begin{adjustbox}{width=1\textwidth}
\begin{tabular}{c|c|c|c|c|c|c|c|c|c|c|c|c|c}
    \hline
      & en & fr & es & de & el & bg & ru & vi& th & zh& hi & ur & avg \\

    \hline

    \hline
    \multicolumn{14}{c}{two languages \cite{nooralahzadeh_zero-shot_2020}   }\\
    \hline
     Multi-BERT (Baseline) & $81.94$& $75.39$&$ 75.79$ &$73.25$ &$69.54$&$71.60$ &$70.84$ &$73.23$& $61.18$&$73.93$ &$64.37$ &$63.71$ &$71.23$\\
     XMAML  & $82.71$ & $75.97$ & $76.51$ & $74.07$ & $70.66$ & $72.77$ & $72.12$ & $73.87$ & \boldmath$62.5$ & $74.85$ & $65.75$ & $64.59$ & $72.20$\\ 
     \hline
     \hline
    \multicolumn{14}{c}{all languages (ours)}\\
    \hline
     Multi-BERT (Baseline) & $83.56$ & $76.22$ & $76.89$ & $73.11$ & \boldmath$72.89$ & $72.89$ & $71.33$ & $74.67$ & $57.56$ & $74.89$ & $63.11$ & $63.33$ & $71.70$ \\
    MAML & $83.11$ & $78.22$ & $77.11$ & $73.56$ & $69.33$ & $71.78$ & $71.33$ & $74.22$ & $57.33$ & $75.11$ & $63.33$ & $63.78$ & $71.52$ \\
     \textbf{Fixed TreeMAML } & \boldmath$84.67$ & \boldmath$79.78$ & $78.22$ & $76.89$ & $72.00$ & \boldmath$74.22$ & $73.33$ & $74.44$ & $59.56$ & \boldmath$79.11$ &\boldmath $66.00$ & \boldmath$66.89$ & \boldmath$73.76$ \\
     \textbf{Learned TreeMAML} & $84.22$ & $77.33$ & \boldmath$79.78$ & \boldmath$78.00$ & $71.56$ & $73.78$ & \boldmath$74.00$ & \boldmath$74.89$ & $59.78$ & $76.44$ & $65.11$ & $65.56$ & $73.37$ \\

\end{tabular}
\end{adjustbox}
\caption{\label{tab: NLIResults}The top part of the table shows the results of training with two auxiliary languages \cite{nooralahzadeh_zero-shot_2020}. The lower part of the table shows the performance when using all languages (except the target) as auxiliary languages. The difference in the amount of data used for training may account for a part of the difference in performance between XMAML and TreeMAML, and may also explain why our Baseline outperforms XMAML for some languages. The results are reported for each of our experiment by averaging the performance over three different runs. The standard deviation is for all our experiments below 1\%.}
\end{table*}
As shown in Table \ref{tab: NLIResults}, fixed/learned TreeMAML outperforms other methods in almost all languages. These results show that using the languages hierarchical structure helps achieving better cross-lingual transfer and higher accuracy in the XNLI task. 

In the case of Greek (el), TreeMAML outperforms XMAML, but the baseline Multi-BERT obtains a slightly higher accuracy. This result could be due to the simplified structure of the tree that we use, which does not adequately reflect the actual distance in between languages from the Indo-European family.
Besides Greek, Thai (th) is the only language for which TreeMAML does not get higher accuracy.  This is mainly due to oversimplified tree used. We used a generic "Asian" language tree, but  Chinese, Vietnamese and Thai belong to three separate language families.

Learned TreeMAML performs very similar to fixed TreeMAML in most experiments, achieving higher values for some of the languages. We believe that the difference depends on how well our clustering algorithm performs in each case.  For some languages, learned TreeMAML is just learning the same tree structure that we use in fixed TreeMAML, and both algorithms produce almost the same results. In other cases, the clustering algorithm assigns tasks to the wrong branch, making learned TreeMAML perform worse. For some other languages, learned TreeMAML performs better, possibly because it finds other useful relationships, for example tasks belonging to the same genre in different languages.

\section{Discussion and Conclusion}
This paper presents a method to exploit the data hierarchy in the meta-learning framework, TreeMAML. This algorithm can use a priory knowledge of the data set (fixed TreeMAML), or learn the hierarchical structure using our modification of the OTD clustering algorithm (learned TreeMAML).

To illustrate the performance and the benefits of this approach, we applied the TreeMAML algorithm to a multidimensional linear regression problem, where the parameters of the tasks are sampled from a distribution that can be described by hierarchical clustering. Both, fixed and learned TreeMAML outperform the baseline and MAML by a significant margin on this synthetic task, halving the MSE of the other methods ( Table \ref{tab: lowNoiseRegResults}). In the case of few shot learning when the number of points of the test task is small, the learned algorithm outperform the fixed tree by exploding similitude between tasks not described by the fixed tree.

Since languages follow a hierarchical phylogenetic tree, we hypothesized that we could use TreeMAML to meta-train models for cross-lingual understanding. We applied TreeMAML to the cross-lingual XNLI problem and show an improvement in accuracy  $\sim3\%$ with respect to the state of the art obtained by XMAML \cite{nooralahzadeh_zero-shot_2020} (Table \ref{tab: NLIResults}). The improvement with respect to XMAML suggests that using all available languages results in increased performance.
Furthermore, the improvement with respect to MAML suggests that using the tree structure of those languages also improves performance.

How much each auxiliary language contributes to the target language's performance may depend on its relative position in the language tree.  These results are especially encouraging for meta-training of cross-lingual understanding tasks for under-resourced languages. 
Future work may include an improved algorithm that takes into account not only the position of a language in the tree, but also the distances between languages in a branch by, for example, introducing weighted averaging of the gradients.  

In our NLI experiments, learned TreeMAML is in most cases as good or even better than fixed TreeMAML. One possible explanation is that clustering learns the tree for each batch of data at each gradient step, which allows it to pick up NLI-relevant similarities that are not described by a phylogenetic tree, as the genre of the task in the XNLI data set, structural similarities and lexical similarity, which can be the result of language contact, as for example lexical borrowings.
This may help with cross-lingual understanding tasks for uncommon languages for which the exact position in the tree may be unclear or not enough data may be available. 

As discussed in section \ref{treemaml-nli}, the lack of improvement in accuracy in the Greek language and the low performance in Thai can be rooted in the experimental design's naive assumptions about: which languages to include in the experiment, and the correctness of the language tree. 
The results in which TreeMAMl performs worse than the other algorithms speak in favour of the robustness of this algorithm to properly learn cross-lingual relationships and exploit them to perform natural language understanding tasks. 
Therefore, the use of learned TreeMAML could help with the internationalization of the NLP models.

\bibliography{kk}
\bibliographystyle{acl_natbib}

\end{document}

%% file: math_commands.tex
\usepackage{amsmath,amsfonts,bm}

\def\eqref#1{equation~\ref{#1}}

\def\1{\bm{1}}

\DeclareMathAlphabet{\mathsfit}{\encodingdefault}{\sfdefault}{m}{sl}
\SetMathAlphabet{\mathsfit}{bold}{\encodingdefault}{\sfdefault}{bx}{n}

\DeclareMathOperator*{\argmin}{arg\,min}

%% file: arxiv_TreeMAML.bbl
\begin{thebibliography}{18}
\expandafter\ifx\csname natexlab\endcsname\relax\def\natexlab#1{#1}\fi

\bibitem[{Achille et~al.(2019)Achille, Lam, Tewari, Ravichandran, Maji,
  Fowlkes, Soatto, and Perona}]{achille_task2vec_2019}
Alessandro Achille, Michael Lam, Rahul Tewari, Avinash Ravichandran, Subhransu
  Maji, Charless Fowlkes, Stefano Soatto, and Pietro Perona. 2019.
\newblock \href {http://arxiv.org/abs/1902.03545} {{Task2Vec}: {Task}
  {Embedding} for {Meta}-{Learning}}.
\newblock \emph{arXiv:1902.03545}.
\newblock ArXiv: 1902.03545.

\bibitem[{Conneau et~al.(2018)Conneau, Lample, Rinott, Williams, Bowman,
  Schwenk, and Stoyanov}]{conneau_xnli_2018}
Alexis Conneau, Guillaume Lample, Ruty Rinott, Adina Williams, Samuel~R.
  Bowman, Holger Schwenk, and Veselin Stoyanov. 2018.
\newblock \href {http://arxiv.org/abs/1809.05053} {{XNLI}: {Evaluating}
  {Cross}-lingual {Sentence} {Representations}}.
\newblock \emph{arXiv:1809.05053}.

\bibitem[{Devlin et~al.(2018)Devlin, Chang, Lee, and
  Toutanova}]{devlin_bert_2018}
Jacob Devlin, Ming-Wei Chang, Kenton Lee, and Kristina Toutanova. 2018.
\newblock \href {http://arxiv.org/abs/1810.04805} {{BERT}: {Pre}-training of
  {Deep} {Bidirectional} {Transformers} for {Language} {Understanding}}.
\newblock \emph{arXiv:1810.04805}.

\bibitem[{Dou et~al.(2019)Dou, Yu, and Anastasopoulos}]{dou_investigating_2019}
Zi-Yi Dou, Keyi Yu, and Antonios Anastasopoulos. 2019.
\newblock \href {http://arxiv.org/abs/1908.10423} {Investigating
  {Meta}-{Learning} {Algorithms} for {Low}-{Resource} {Natural} {Language}
  {Understanding} {Tasks}}.
\newblock \emph{arXiv:1908.10423}.

\bibitem[{Dunn et~al.(2011)Dunn, Greenhill, Levinson, and
  Gray}]{dunn_evolved_2011}
Michael Dunn, Simon~J. Greenhill, Stephen~C. Levinson, and Russell~D. Gray.
  2011.
\newblock \href {https://doi.org/10.1038/nature09923} {Evolved structure of
  language shows lineage-specific trends in word-order universals}.
\newblock \emph{Nature}, 473(7345):79--82.

\bibitem[{Finn et~al.(2017)Finn, Abbeel, and Levine}]{finn_model-agnostic_2017}
Chelsea Finn, Pieter Abbeel, and Sergey Levine. 2017.
\newblock \href {http://arxiv.org/abs/1703.03400} {Model-{Agnostic}
  {Meta}-{Learning} for {Fast} {Adaptation} of {Deep} {Networks}}.
\newblock \emph{arXiv:1703.03400}.
\newblock ArXiv: 1703.03400.

\bibitem[{Gu et~al.(2018)Gu, Wang, Chen, Cho, and Li}]{gu_meta-learning_2018}
Jiatao Gu, Yong Wang, Yun Chen, Kyunghyun Cho, and Victor O.~K. Li. 2018.
\newblock \href {http://arxiv.org/abs/1808.08437} {Meta-{Learning} for
  {Low}-{Resource} {Neural} {Machine} {Translation}}.
\newblock \emph{arXiv:1808.08437}.

\bibitem[{Hospedales et~al.(2020)Hospedales, Antoniou, Micaelli, and
  Storkey}]{hospedales_meta-learning_2020}
Timothy Hospedales, Antreas Antoniou, Paul Micaelli, and Amos Storkey. 2020.
\newblock \href {http://arxiv.org/abs/2004.05439} {Meta-{Learning} in {Neural}
  {Networks}: {A} {Survey}}.
\newblock \emph{arXiv:2004.05439}.
\newblock ArXiv: 2004.05439.

\bibitem[{Jerfel et~al.(2019)Jerfel, Griffiths, Grant, and
  Heller}]{jerfel_reconciling_2019}
Ghassen Jerfel, Thomas~L Griffiths, Erin Grant, and Katherine Heller. 2019.
\newblock Reconciling meta-learning and continual learning with online mixtures
  of tasks.
\newblock \emph{NIPS}, page~12.

\bibitem[{Katoch et~al.(2020)Katoch, Thopalli, Thiagarajan, Turaga, and
  Spanias}]{katoch_invenio_2020}
Sameeksha Katoch, Kowshik Thopalli, Jayaraman~J. Thiagarajan, Pavan Turaga, and
  Andreas Spanias. 2020.
\newblock \href {http://arxiv.org/abs/1911.10600} {Invenio: {Discovering}
  {Hidden} {Relationships} {Between} {Tasks}/{Domains} {Using} {Structured}
  {Meta} {Learning}}.
\newblock \emph{arXiv:1911.10600}.
\newblock ArXiv: 1911.10600.

\bibitem[{Liu et~al.(2019)Liu, Lee, Park, Kim, Yang, Hwang, and
  Yang}]{liu_learning_2019}
Yanbin Liu, Juho Lee, Minseop Park, Saehoon Kim, Eunho Yang, Sung~Ju Hwang, and
  Yi~Yang. 2019.
\newblock \href {http://arxiv.org/abs/1805.10002} {Learning to {Propagate}
  {Labels}: {Transductive} {Propagation} {Network} for {Few}-shot {Learning}}.
\newblock \emph{arXiv:1805.10002}.
\newblock ArXiv: 1805.10002.

\bibitem[{Menon et~al.(2019)Menon, Rajagopalan, Sumengen, Citovsky, Cao, and
  Kumar}]{menon_online_2019}
Aditya~Krishna Menon, Anand Rajagopalan, Baris Sumengen, Gui Citovsky, Qin Cao,
  and Sanjiv Kumar. 2019.
\newblock \href {http://arxiv.org/abs/1909.09667} {Online {Hierarchical}
  {Clustering} {Approximations}}.
\newblock \emph{arXiv:1909.09667}.
\newblock ArXiv: 1909.09667.

\bibitem[{Nooralahzadeh et~al.(2020)Nooralahzadeh, Bekoulis, Bjerva, and
  Augenstein}]{nooralahzadeh_zero-shot_2020}
Farhad Nooralahzadeh, Giannis Bekoulis, Johannes Bjerva, and Isabelle
  Augenstein. 2020.
\newblock \href {http://arxiv.org/abs/2003.02739} {Zero-{Shot}
  {Cross}-{Lingual} {Transfer} with {Meta} {Learning}}.
\newblock \emph{arXiv:2003.02739}.
\newblock ArXiv: 2003.02739.

\bibitem[{Ruder(2017)}]{ruder_overview_2017}
Sebastian Ruder. 2017.
\newblock \href {http://arxiv.org/abs/1706.05098} {An {Overview} of
  {Multi}-{Task} {Learning} in {Deep} {Neural} {Networks}}.
\newblock \emph{arXiv:1706.05098}.
\newblock ArXiv: 1706.05098.

\bibitem[{Srivastava and Salakhutdinov(2013)}]{srivastava_discriminative_2013}
Nitish Srivastava and Russ~R Salakhutdinov. 2013.
\newblock \href
  {http://papers.nips.cc/paper/5029-discriminative-transfer-learning-with-tree-based-priors.pdf}
  {Discriminative {Transfer} {Learning} with {Tree}-based {Priors}}.
\newblock In C.~J.~C. Burges, L.~Bottou, M.~Welling, Z.~Ghahramani, and K.~Q.
  Weinberger, editors, \emph{Advances in {Neural} {Information} {Processing}
  {Systems} 26}, pages 2094--2102. Curran Associates, Inc.

\bibitem[{Yao et~al.(2019)Yao, Wei, Huang, and Li}]{yao_hierarchically_2019}
Huaxiu Yao, Ying Wei, Junzhou Huang, and Zhenhui Li. 2019.
\newblock \href {http://arxiv.org/abs/1905.05301} {Hierarchically {Structured}
  {Meta}-learning}.
\newblock \emph{arXiv:1905.05301}.
\newblock ArXiv: 1905.05301.

\bibitem[{Zamir et~al.(2018)Zamir, Sax, Shen, Guibas, Malik, and
  Savarese}]{zamir_taskonomy_2018}
Amir Zamir, Alexander Sax, William Shen, Leonidas Guibas, Jitendra Malik, and
  Silvio Savarese. 2018.
\newblock \href {http://arxiv.org/abs/1804.08328} {Taskonomy: {Disentangling}
  {Task} {Transfer} {Learning}}.
\newblock \emph{arXiv:1804.08328}.
\newblock ArXiv: 1804.08328.

\bibitem[{Zhang and Yang(2018)}]{zhang_survey_2018}
Yu~Zhang and Qiang Yang. 2018.
\newblock \href {http://arxiv.org/abs/1707.08114} {A {Survey} on {Multi}-{Task}
  {Learning}}.
\newblock \emph{arXiv:1707.08114}.
\newblock ArXiv: 1707.08114.

\end{thebibliography}
